\documentclass{bmvc2k}

\title{Decentralised Person Re-Identification with Selective Knowledge Aggregation}

\addauthor{Shitong Sun}{shitong.sun@qmul.ac.uk}{1}
\addauthor{Guile Wu}{guile.wu@qmul.ac.uk}{1}
\addauthor{Shaogang Gong}{s.gong@qmul.ac.uk}{1}

\addinstitution{Computer Vision Group,\\
School of Electronic Engineering and \\
Computer Science,\\
Queen Mary University of London,\\
London E1 4NS, UK}

\runninghead{S. Sun, G. Wu, S. Gong}{SELECTIVE KNOWLEDGE AGGREGATION}

\usepackage{lipsum}
\usepackage{algorithm}
\usepackage{multicol}
\usepackage[noend]{algpseudocode}

\usepackage{amssymb}
\usepackage{graphicx}
\usepackage{multirow}

\usepackage{tabularx}
\usepackage{arydshln}

\usepackage{multicol}

\def\etal{\emph{et al}\bmvaOneDot}

\def\ie{\emph{i.e}\bmvaOneDot}

\begin{document}

\maketitle

\begin{abstract}
    Existing person re-identification (Re-ID) methods mostly follow a centralised learning paradigm
    which shares all training data to a collection for model learning.
    This paradigm is limited when data from different sources cannot be shared due to privacy concerns.
    To resolve this problem, two recent works~\cite{wu2021decentralised,zhuang2020performance} have
    introduced decentralised (federated) Re-ID learning for constructing a
    globally generalised model (server) without any direct access to local training data
    nor shared data across different source domains (clients).
    However, these methods are poor on how to adapt
    the generalised model to maximise its performance on individual
    client domain Re-ID tasks
    having different Re-ID label spaces, due to a lack
    of understanding of data heterogeneity across domains.
    We call this poor `model personalisation'.
    In this work, we present a new Selective Knowledge Aggregation approach to decentralised person
    Re-ID to optimise the trade-off between model personalisation and generalisation.
    Specifically, we incorporate attentive normalisation into the normalisation layers in a deep ReID model
    and propose to learn local normalisation layers specific to each domain,
    which are decoupled from the global model aggregation in federated Re-ID learning.
    This helps to preserve model personalisation knowledge on each local client domain
    and learn instance-specific information.
    Further, we introduce a dual local normalisation mechanism to learn generalised
    normalisation layers in each local model, which are then transmitted to the global model for central aggregation.
    This facilitates selective knowledge aggregation on the server to construct
    a global generalised model for out-of-the-box deployment on unseen novel domains.
    Extensive experiments on eight person Re-ID datasets show that
    the proposed approach to decentralised Re-ID significantly outperforms the state-of-the-art
    decentralised methods on both seen client domains and unseen novel domains.
\end{abstract}

\section{Introduction}
\label{sec:intro}
Person re-identification (Re-ID) attempts to retrieve a person of interest from a set of gallery images captured
from non-overlapping camera views~\cite{ye2021deep,jin2020style,wu2020tracklet,li2018harmonious,xu2018attention}.
It plays an important role in a wide range of real-world applications,
such as finding a missing person, smart city management. 
Existing person Re-ID methods mostly follow a centralised learning paradigm
which requires to collect all training data from different camera views or domains together for model learning.
Despite significant progress has been made, centralised Re-ID learning
ignores that person images contain a large amount of personal privacy information
which may not be allowed to be shared to a central data collection.
This limits fundamentally existing centralised learning based Re-ID in real-world
applications with increasing privacy-sensitive scenarios.

To solve this problem, two recent Re-ID works~\cite{wu2021decentralised,zhuang2020performance}
have applied federated learning~\cite{mcmahan2017communication} to person re-identification
by constructing a global generalised model (server) without
access to local training data nor sharing data across different source domains (clients).
This privacy-preserving Re-ID learning paradigm is known as
{\em decentralised person Re-ID} \cite{wu2021decentralised}.
More specifically, in decentralised person Re-ID,
each local client trains its local model using its own set of local training data
without sharing data with other local clients,
whilst a central server generates a global model by aggregating local model weights
(or local model updates) without any direct access to training data.
This decentralised person Re-ID paradigm inherently protects source data privacy.
Although latest decentralised Re-ID methods
\cite{wu2021decentralised,zhuang2020performance} have shown
encouraging performance, their focus is on how to learn a global generalised model
and they have ignored the need to understand data heterogeneity across domains.
As a result, they are sub-optimal when compared to 
local trained models for individual client domain Re-ID tasks having different Re-ID label
spaces, \ie showing poor `model personalisation' on each local client domain.

In this work, we propose to optimise the trade-off between model personalisation and generalisation
in decentralised person Re-ID.
We present a new Selective Knowledge Aggregation (SKA) approach to facilitate feature extraction model
learning in the iterative federated Re-ID learning paradigm~\cite{wu2021decentralised,zhuang2020performance}.
Specifically, we incorporate attentive normalisation~\cite{li2019attentive} into the normalisation layers
in a deep ReID model and propose to learn local normalisation layers specific to each client.
These local normalisation layers are decoupled from the central server
aggregation whilst the other
layers in the feature extraction models are iteratively updated between the clients and the server.
Learning these local specific normalisation layers helps to better represent model personalisation knowledge
from instance-specific information for tuning
`downwards' the server to each participating client domain. 
Crucially, these local client specific normalisation layers cannot be
aggregated `upwards' to the central server,
because directly updating them iteratively to the central server, 
e.g. by averaging as in
existing decentralised Re-ID models, will degrade server
generalisation to unseen novel domains.
To compensate this emission of direct upwards aggregation to the
central server, we introduce a dual local normalisation mechanism to additionally
learn generalised normalisation layers in each local client,
which are iteratively updated between the clients and the server.
This improves to construct a globally generalised model for better
out-of-the-box deployment on unseen novel domains. 

The key {\bf\emph{contributions}} of this work are:
We propose a new Selective Knowledge Aggregation approach to optimise the trade-off between model
personalisation and generalisation in decentralised person Re-ID.
To learn model personalisation knowledge from instance-specific information per local client,
we incorporate attentive normalisation into the normalisation layers in a deep Re-ID model
so to learn local normalisation layers specific to each client domain.
To replace the direct global model aggregation in existing
decentralised learning methods,
we introduce a dual local normalisation mechanism which additionally learns
generalised normalisation layers in each local client domain
for constructing a better global generalised model for unseen novel domains.

Extensive experiments on eight person Re-ID datasets, including
DukeMTMC-ReID~\cite{zheng2017unlabeled}, Market-1501~\cite{zheng2015scalable}, 
CUHK03-NP~\cite{li2014deepreid,zhong2017re}, MSMT17~\cite{wei2018person}, VIPeR~\cite{gray2008viewpoint},
iLIDS~\cite{zheng2009associating}, GRID~\cite{loy2013person} and PRID~\cite{hirzer11},
show that the proposed SKA model significantly outperforms
the state-of-the-art decentralised methods on both seen client domains and unseen novel domains.

\section{Related Work}
\noindent\textbf{Person Re-Identification} (Re-ID) research has been very
active in computer vision in recent years. 
It aims to match a person of interest across non-overlapping camera
views~\cite{ye2021deep,jin2020style,wu2020tracklet,wu2019spatio}.
Traditional person Re-ID methods~\cite{li2018harmonious,xu2018attention,zheng2015scalable}
mainly follow a supervised learning approach,
which requires to collect all training data from different camera views in the target domain for model training
and usually shows poor generalisation to an unseen novel domain.
Recently, generalisable person Re-ID~\cite{song2019generalizable,jin2020style,choi2021meta},
which attempts to generalise a model with data from different domains,
has attracted increasing attention and shown promising performance.
For example, Song~\etal~\cite{song2019generalizable} assemble data from multiple domains to
optimise a domain-invariant mapping network for out-of-the-box Re-ID deployment.
Choi~\etal~\cite{choi2021meta} propose to generalise batch-instance normalisation layers
in a meta-learning pipeline for learning a generalisable Re-ID model.
However, these generalisable person Re-ID methods require to collect training data from different domains
together to learn a feature extraction model for person matching.
This largely ignores data privacy and is inherently limited in
situations where data from different domains cannot be shared to a central data collection.
To resolve this problem, two recent works~\cite{wu2021decentralised,zhuang2020performance}
propose an iterative federated Re-ID learning paradigm to learn a global generalised model with
the collaboration of local models but without sharing local data.
This decentralised person Re-ID paradigm can inherently protect local data privacy
whilst generalising a Re-ID model for deployment.
However, current decentralised methods
\cite{wu2021decentralised,zhuang2020performance} fail to consider how
to adapt the generalised model 
to maximise the performance on individual domain Re-ID tasks.
Our work focuses on decentralised person Re-ID model learning from multiple domains
without sharing local data across domains so as to protect data privacy.
But different from~\cite{wu2021decentralised,zhuang2020performance}, we propose to optimise the trade-off
between model personalisation and generalisation by learning local normalisation layers specific to each domain
and employ a dual local normalisation mechanism to facilitate
aggregating a better global generalised model for both client domains
and unseen novel domains.

\vspace{0.1cm}
\noindent\textbf{Federated Learning} aims to learn a global model with the collaboration of multiple local models.
In the seminal paper~\cite{mcmahan2017communication}, McMahan~\etal introduce a FedAvg algorithm to federated learning,
which iteratively averages local model updates to construct a global model.
Inspired by the success of FedAvg,
there have been many promising works~\cite{wang2020federated,li2021fedbn,he2020group,wang2020tackling,li2018federated,shen2020federated,peng2019federated,liang2020think,li2021model,lin2020ensemble}
that present more effective federated learning algorithms.
Due to non-IID (Independent and Identically Distributed) data distribution across clients,
a global generalised model usually shows poor personalisation performance to each specific client.
Recently, some works have introduced personalised federated
learning~\cite{tan2021towards,t2020personalized,karimireddy2020scaffold}
which aims to improve model personalisation specific to each client with non-IID data across clients.
SCAFFOLD~\cite{karimireddy2020scaffold} estimates the update direction of the global model to further
regularise the update direction of the local client, so as to improve local model personalisation.
PFedMe~\cite{t2020personalized} employs Moreau envelopes as the regularised loss for local model optimisation
to facilitate personalised model learning.
Our work focuses on optimising the trade-off between model personalisation and
generalisation for decentralised person Re-ID.
Different from contemporary federated learning,
our proposed SKA method incorporates attentive normalisation into the normalisation layers
and optimises local normalisation layers in a Re-ID model to explore model personalisation knowledge
from instance-specific information.
Moreover, a dual local normalisation mechanism is introduced to
optimise global model generalisation to unseen novel domains.

\begin{figure}[t]
    \centering
    \includegraphics[width=1.0\textwidth]{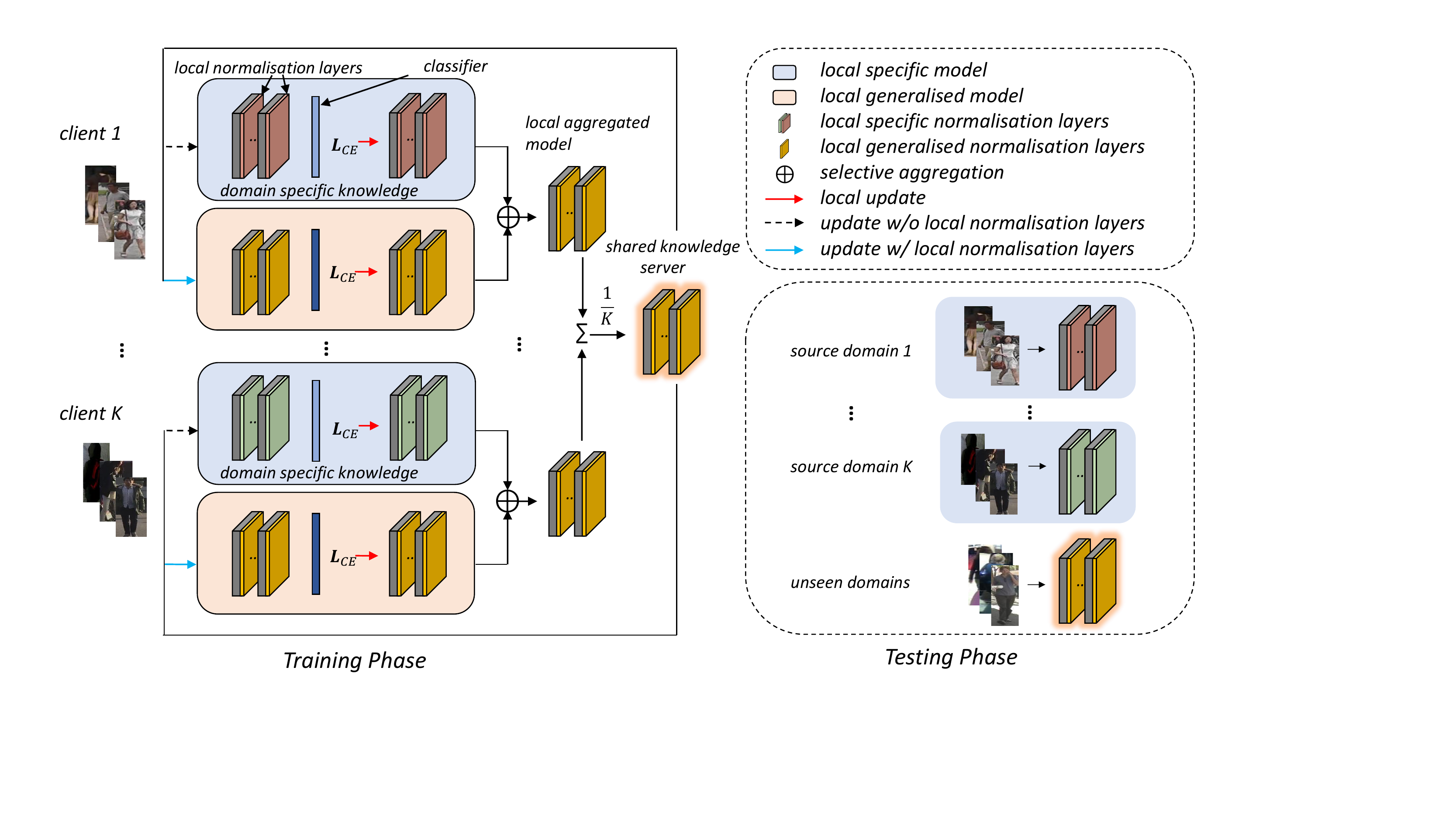}
    \caption{
    An overview of the Selective Knowledge Aggregation (SKA) method
    for decentralised person Re-ID.
    In each training domain, a local client model is optimised to learn local specific normalisation layers
    (layers in pink and green)
    for improving \emph{model personalisation}, whilst a local
    generalised model is optimised to learn local generalisable normalisation layers (layers in yellow)
    for improving \emph{model generalisation} globally.
    On the central server, a global generalised model is both
    aggregated centrally and then used to update client models locally.
    }
     \label{fig:overview}
 \end{figure}

\section{Methodology}
\subsection{Overview}
In this work, we study decentralised person Re-ID,
where data from different domains (clients) are private and the global model (server)
can only aggregate local models to learn generalised knowledge across domains.
The aim of the proposed SKA method is to optimise the trade-off between model personalisation and
generalisation in decentralised person Re-ID.
Figure~\ref{fig:overview} shows an overview of the proposed SKA method.

Suppose there are $N$ local client domains where each client contains a local non-shared private dataset.
As shown in Figure~\ref{fig:overview},
each local client builds two local models concurrently: A local specific model
(the block in blue) and a local generalised model (the block in orange).
The local specific model learns local normalisation layers (layers in pink and green)
specific to each domain which are decoupled from the
central aggregation whilst the other layers (layers in grey) in this model
are transmitted to the server for central aggregation.
In parallel, the local generalised model learns generalised
normalisation layers (layers in yellow) which are all transmitted to
the server for central aggregation.
These two local models are trained simultaneously (for $E$ local epochs) using a cross-entropy loss.
Afterwards, a selective aggregation is conducted and generates an aggregated local model.
Next, on the central server, a global generalised model is constructed with the collaboration of aggregated local models
and then is transmitted to each local client to further facilitate local model learning.
The local generalised models update all layers in the feature extraction network with the global generalised model,
whilst the local specific models retain local normalisation layers but update other layers in the feature extraction network.
Note, following~\cite{wu2021decentralised,zhuang2020performance}, local domain-specific classifiers are decoupled from
the global model aggregation. This differs from the contemporary FedAvg~\cite{mcmahan2017communication}
that aggregates all model components.
After $T$ global iterations of federated Re-ID model learning,
a global generalised feature model is learnt on the server for
out-of-the-box Re-ID deployments to unseen novel domains. 
Each local client model with local specific normalisation layers can
then be combined with this global model for better model
personalisation in Re-ID deployment to each client domain.

\subsection{Learning Local Model Personalisation Knowledge}
\label{sec:LPK}
Existing decentralised person Re-ID works~\cite{wu2021decentralised,zhuang2020performance}
aggregate all layers of a feature representation network to construct a global generalised model
and fail to tackle data heterogeneity across clients (domains),
resulting in sub-optimal performance on participating client domains.
To solve this problem, we propose to learn local `personalisation' knowledge in each client domain
to maximise model performance per domain Re-ID tasks,
and this local personalisation knowledge is encoded in local normalisation
layers of a deep learning network.

Specifically, batch normalisation (BN)~\cite{ioffe2015batch} layers in a deep network
are prevailingly used to normalise activations using data in a mini-batch
which captures information about the data distribution~\cite{li2021fedbn}.
Recently, some works~\cite{chang2019domain,seo2019learning,li2021fedbn} have shown that
learning domain-specific batch normalisation (BN)~\cite{ioffe2015batch} layers is beneficial to better tackle the domain shift problem across domains.
In the context of federated Re-ID model learning,
person images from different client domains are almost entirely in
non-overlapping label spaces,
which exacerbates the domain shift problem.
Thus, to preserve personalisation knowledge on each participating client domain
and resolve the domain shift problem,
we decouple normalisation layers in each local network from central aggregation
and optimise local normalisation layers specific to each client.

However, only optimising the vanilla BN layers for each client can lead to sub-optimal performance
because for each local client, although person images are from the same domain,
there are usually notorious misalignment of person images caused by pose change,
imperfect detection, occlusion, significant variation in lighting.
Thus, it is important to learn instance-specific information on each client domain
whilst tackling data heterogeneity.
Attention modelling is an effective solution to address this problem.
To learn instance-specific channel-wise attention weights to re-calibrate features,
the Squeeze-and-Excitation (SE) attention unit~\cite{hu2018squeeze} can be incorporated into a deep network model
as follows:
\begin{equation}
    \label{eq:bnse}
    \tilde{x}^{S E}=\lambda\cdot {BN}(x)=
    \left(\lambda \cdot \gamma\right) \cdot \hat{x}+\lambda\cdot \beta
\end{equation}
where ${BN}(x)=\gamma\left(\frac{x-\mu(x)}{\sigma(x)}\right)+\beta$, $\mu(x), \sigma(x)$ are the mean and standard 
deviation computed across batch size and spatial dimensions independently for each feature channel, $\hat{x} $ is the
normalised feature, $\lambda $ is the learnt attentive weight,
$\gamma, \beta $ are affine parameters.
Since affine transform parameters in BN and the re-scaling parameters of the SE unit
are both learning to re-calibrate features,
following~\cite{li2019attentive},
they can be further combined as $M$ mixture components of channel-wise affine transform:
\begin{equation}
    \label{eq:AN}
    \tilde{x}^{A N}=\sum_{i=1}^{M} \lambda_{i}\left[\gamma_{i} \cdot \hat{x}+\beta_{i}\right]
\end{equation}
where $\lambda_{i} $, $\gamma_{i}, \beta_{i} $ are the $i$-th mixture integrated into a compact module and learnt simultaneously,
so they can learn more shared instance-specific information.
We therefore combine some BN layers and attention modelling in a deep
network for each local client model and decouple them from global server aggregation.
This approach to learning local model personalisation knowledge brings
significant benefits to model performance (see Section~\ref{section_ablation}).

\begin{algorithm}[t]
  \footnotesize
      \caption{\textbf{Decentralised Re-ID with Selective Knowledge Aggregation (SKA)}.
      The $K$ clients are indexed by $k$, $\mathcal{N}_{k}$ indicates the dataset on client $k$,
      $E$ is the number of local epochs, the number of global iteration rounds is $T$, $\eta$ is the learning rate,
      $w$ are the model weights,
      where $w_{ls}$, $w_{lg}$ and $w_{l}$ refer to model weights of a local specific model, a local generalised model
      and a local aggregated model respectively.
      Steps 1-8 are the main process, while steps 10-15, 17-23 and 25-31 are functions.
    }

    \label{algo:SKA}

    \begin{multicols}{2}
\begin{algorithmic}[1]
    \State \textbf{initialize} \{$w^k_{0,ls}$,$w^k_{0,lg}\}_{k=1}^K$
    \For{global round $t$ = 0 to $T-1$} 
    \For{each client $k = 1$ to $K$}
    \State $w^{k}_{t+1,ls},w^{k}_{t+1,lg} \leftarrow$ \textbf{LocalUpdate}($w^{k}_{t,ls},w^{k}_{t,lg}$)
    \State $w^{k}_{t+1,l} \leftarrow $ \textbf{SelectAgg}($w^k_{t+1,ls},w^k_{t+1,lg}$)
    \EndFor
    \State $w_{t+1} = \frac{1}{K}\sum_{k=1}^{K} w^k_{t+1,l}$
    \State \{$w^{k}_{t+1,ls},w^{k}_{t+1,lg}\}_{k=1}^K \leftarrow$ \textbf{SelectUpdate}($w_{t+1}$)
    \EndFor
    \State\Return $\{w^k_{T,ls}\}_{k=1}^K,w_{T}$
    \\
    \Function{LocalUpdate}{$w^{k}_{t,ls},w^{k}_{t,lg}$}
    \For{local epoch $e$ = 0 to $E-1$}
    \For{mini-batch $b \subset \mathcal{N}_{k}$}
    \State  $w^{k}_{t+1,ls} \leftarrow   w^k_{t,ls}-\eta \nabla \ell(w^k_{t,ls} )$ 
    \State $w^{k}_{t+1,lg} \leftarrow  w^k_{t,lg}-\eta \nabla \ell(w^k_{t,lg} )$ 
    \EndFor
    \EndFor
    \State\Return $w^{k}_{t+1,ls},w^{k}_{t+1,lg}$
    \EndFunction
    \\
    \Function{SelectAgg}{$w^k_{t+1,ls},w^k_{t+1,lg}$}
    \For{layer $p$ in $w_{t+1,ls}^k$}
    \If{layer $p$ is normalisation layer}
    \State $w_{t+1,l,p}^k \leftarrow w_{t+1,lg,p}^k$
    \Else
    \State $w_{t+1,l,p}^k \leftarrow w_{t+1,ls,p}^k$
    \EndIf
    \EndFor
    \State\Return $w_{t+1,l}^k$
    \EndFunction
    \\
    \Function{SelectUpdate}{$w_{t+1}$}
    \For{each client $k$}
    \State $w_{t+1,lg}^k \leftarrow w_{t+1}$
    \For{layer $p$ in $w_{t+1}$}
    \If{$p$ is NOT normalisation layer}
    \State $ w_{t+1,ls,p}^k \leftarrow w_{t+1,p} $      
    \EndIf
    \EndFor
    \EndFor
    \State\Return \{$w_{t+1,ls}^k,w_{t+1,lg}^k \}_{k=1}^K$
    \EndFunction
\end{algorithmic}
\end{multicols}
\end{algorithm}

\vspace{0.1cm}
\subsection{Dual Local Normalisation for Global Generalisation Knowledge}

\label{sec:dual}
Although decoupling normalisation layers from central aggregation helps to optimise
local personalisation knowledge,
it will also lose some generalisable knowledge useful in constructing
a better global model for unseen novel domains.
A simple solution is to average local specific normalisation layers at the end of model training,
but this can lead to sub-optimal global generalisation because
local normalisation layers are not aggregated in the iterative global
model update.
To compensate this degradation in the global model learning,
we introduce a dual local normalisation mechanism to explicitly learn
auxiliary local generalised normalisation layers.

As shown in Figure~\ref{fig:overview}, in addition to the local specific model,
we optimise a local generalised model (with the same network architecture) to learn local generalised normalisation layers.
Then, for each client, we use local generalised normalisation layers (layers in yellow in Figure~\ref{fig:overview})
from a local generalised model and other layers (layers in grey in Figure~\ref{fig:overview})
from a corresponding local specific model to construct a local aggregated model.
The local aggregated models from different clients are then transmitted to the server for iterative central aggregation.
This helps to alleviate the degradation problem because the generalised normalisation layers
are iteratively updated between the client and the server and are not specific to a certain client.
As a result, the global model cumulates richer generalised knowledge from local
clients that can benefit significantly model performance on unseen
novel domains (see Table~\ref{tab:sota_small} in Section~\ref{section_SOTA}).
Besides, since the dual local normalisation mechanism compensate the degradation brought by learning local
personalisation knowledge, it facilitates the proposed SKA to make a trade-off between learning
model personalisation and generalisation.
We summarise the model training process of the proposed SKA method in Algorithm~\ref{algo:SKA}.

\section{Experiments}
\subsection{Datasets and Evaluation Protocol}
\noindent\textbf{Datasets}:
To evaluate the proposed SKA model,
we used four large-scale person Re-ID datasets
Market1501~\cite{zheng2015scalable}, DukeMTMC-ReID~\cite{zheng2017unlabeled},
CUHK03-NP~\cite{li2014deepreid,zhong2017re} and MSMT17~\cite{wei2018person}
as local client domains.
Data on each client are only used for local model training without sharing to other clients nor the server.
Furthermore, we employed four smaller person Re-ID datasets
VIPeR~\cite{gray2008viewpoint}, iLIDS~\cite{zheng2009associating},
GRID~\cite{loy2013person} and PRID~\cite{hirzer11} 
as unseen novel domains to evaluate the generalisation performance of the global model.
Following~\cite{song2019generalizable,wu2021decentralised},
we generated ten random training/testing splits on each smaller dataset,
and on each test split,
one image of each person identity was used as the query whilst the other images were used as the gallery.
Table~\ref{tab:datasets} summarises the dataset statistics.

\vspace{0.1cm}
\noindent\textbf{Evaluation Metrics}:
We used Rank-1 (R1) accuracy and mean Average Precision (mAP) for Re-ID performance evaluation.

\newcommand{\tabincell}[2]{\begin{tabular}{@{}#1@{}}#2\end{tabular}}
\begin{table}[t]
  \small	
  \begin{center}
    \resizebox{0.9\textwidth}{!}{%
    \begin{tabular}{clccccc}
    \hline
    \multirow{2}{*}{Types} &
      \multicolumn{1}{c}{\multirow{2}{*}{Datasets}} &
      \multirow{2}{*}{Train ID} &
      \multirow{2}{*}{Train Img} &
      \multicolumn{3}{c}{Test} \\ \cline{5-7} 
     &
      \multicolumn{1}{c}{} &
       &
       &
      \multicolumn{1}{l}{Test ID} &
      \multicolumn{1}{l}{Query Img} &
      \multicolumn{1}{l}{Gallery Img} \\ \hline
    \multirow{4}{*}{{\tabincell{c}{Source\\client\\domains}}}
                                        & Market1501 & 751  & 12936 & 750  & 3368  & 19732 \\
                                        & DukeMTMC & 702  & 16522 & 702  & 2228  & 17661 \\
                                        & CUHK03-NP  & 767  & 7365  & 700  & 1400  & 5328  \\
                                        & MSMT17     & 1041 & 30248 & 3060 & 11659 & 82161 \\ \hline
    \multirow{4}{*}{{\tabincell{c}{Unseen\\novel\\domains}}}
                                        & VIPeR      & -    & -     & 316  & 316   & 316   \\
                                        & iLIDS      & -    & -     & 60   & 60    & 60    \\
                                        & GRID       & -    & -     & 125  & 125   & 900   \\
                                        & PRID       & -    & -     & 100  & 100   & 649   \\ \hline
    \end{tabular}%
    }
    \caption{Statistics of eight person Re-ID benchmark datasets.}
    \label{tab:datasets}

\end{center}
\end{table}

\subsection{Implementation Details}
Following~\cite{wu2021decentralised,zhuang2020performance},
we employed ResNet-50~\cite{he2016deep} pretrained on ImageNet~\cite{deng2009imagenet}
as the feature extraction network and used two fully-connected layers
as feature vectors for classification.
We used attentive normalisation~\cite{li2019attentive} (pretrained on
ImageNet) to replace the second vanilla batch normalisation layer on each bottleneck block.
In each local client, we used SGD optimiser with Nesterov momentum 0.9 and weight decay 5e-4
and set the learning rate to 0.01 for the feature extraction network and 0.1 for the classifier,
which were decayed by 0.1 every 40 global epochs.
We empirically set batch size to 32, maximum global iterations $T=100$, maximum local optimisation epoch $E=1$.
For fair comparison, all the reproduced methods including local supervised baseline are based on ResNet-50 backbone
following same training process including optimiser, learning rate, global iteration rounds and local epochs.
All of the experiments are implemented in Python and PyTorch.

\begin{table}[t]
  \small    
    \begin{center}
        \resizebox{0.99\textwidth}{!}{%
        \begin{tabular}{c|c|cc|cc|cc|cc}
        \hline
        \multirow{2}{*}{Method} &
          \multirow{2}{*}{Source} &
          \multicolumn{2}{c|}{\textbf{Duke}} &
          \multicolumn{2}{c|}{\textbf{Market1501}} &
          \multicolumn{2}{c|}{\textbf{CUHK03-NP}} &
          \multicolumn{2}{c}{\textbf{MSMT17}} \\ \cline{3-10} 
           &           &    mAP  &  R1     & mAP &. R1   &  mAP  & R1     & mAP   & R1     \\ \hline
        local supervised &        & 57.0   & 77.2  & 68.2 & 87.4 & 39.2  & 43.2  & 28.8 & 60.0    \\\hline
        FedAvg\cite{mcmahan2017communication}+D & AISTATS'17  & 50.9 & 70.8 & 56.6 & 81.3  & 25.2& 27.9  & 25.9 & 55.0  \\
        FedProx\cite{li2018federated}+D  & MLSys'20 & 53.5 & 72.0  & 59.7 & 84.5  & 27.8 & 30.8 & 28.1 & 57.7  \\
        FedPav\cite{zhuang2020performance}    & ACMMM'20  & 51.9 & 71.1 & 53.5 & 78.7 & 23.0 & 26.0 & 26.1 & 54.4  \\
        FedPav+AddData\cite{zhuang2020performance}          & ACMMM'20   & 60.6& 78.4 & 58.0& 82.4  & 26.8  & 29.9& 27.0 & 55.7 \\
        FedReID\cite{wu2021decentralised}          & AAAI'21  & 52.1& 68.0    & 60.1  & 80.2   & -     & -     & -  & 48.4     \\        
        MOON~\cite{li2021model}+D &  CVPR'21   & 53.1&  72.7 & 58.1 & 83.6 & 26.4 & 28.5 & 27.2& 56.6  \\
        FedBN\cite{li2021fedbn}+D    & ICLR'21& 62.8  & 80.0 & 73.1& 90.4 & 40.9 & 45.3 & 35.4  & 67.6 \\  \hdashline
        SKA           & Ours    & \textbf{66.6}  & \textbf{83.7} & \textbf{78.2} & \textbf{92.7}  & \textbf{48.1} & \textbf{53.0}  & \textbf{42.9}& \textbf{73.8} \\ \hline
        \end{tabular}%
        }
        \caption{Comparisons of decentralised learning Re-ID on four source client domains.
        Note that the conventional federated learning methods cannot be directly used for decentralised Re-ID,
        so `+D' means we implement them by decoupling domain-specific classifiers from central aggregation.
        `FedPav+AddData' means using additional central unlabelled data in FedPav for knowledge distillation and weight adjustment.
        `FedAvg+D' and `FedPav' share the same principle but with different hyper-parameters.
        }
    
        \label{tab:sota_large}
    
    \end{center}
    \end{table}

    \begin{table}[t]
      \small    
    \begin{center}
        \resizebox{0.99\textwidth}{!}{%
        \begin{tabular}{c|c|cc|cc|cc|cc}
            \hline
            \multirow{2}{*}{Method} &
            \multirow{2}{*}{Source} &
            \multicolumn{2}{c|}{\textbf{VIPeR}} &
            \multicolumn{2}{c|}{\textbf{iLIDS}} &
            \multicolumn{2}{c|}{\textbf{GRID}} &
            \multicolumn{2}{c}{\textbf{PRID}} \\ \cline{3-10} 
            &         & mAP   & R1    & mAP   & R1    & mAP  & R1    & mAP   & R1       \\ \hline
        FedAvg\cite{mcmahan2017communication}+D   & AISTATS'17     & 48.2   & 44.3  & 73.3  & 69.3 & 24.3 & 20.1  & 19.6  & 15.8 \\
        FedProx\cite{li2018federated}+D  & MLSys'20 & 47.3 & 43.2  & 74.9 & 71.1 &29.1 & 24.8 &31.2  &26.8  \\
        FedPav\cite{zhuang2020performance}    & ACMMM'20 & 49.5 & 44.9 & 72.8 & 68.8  & 25.5 & 21.7 & 37.0 & 31.9  \\
        FedPav+AddData\cite{zhuang2020performance}      & ACMMM'20       & 49.6 & 45.3& 73.1 & 69.0  & 28.7 &24.2 &  34.4 & 28.5  \\
        FedReID\cite{wu2021decentralised}     & AAAI'21       & -     & 46.2  & -     & 69.7  & -   & 24.2     &  -     & -     \\
        MOON~\cite{li2021model}+D &  CVPR'21 & 49.1 &  45.1 & 73.7 & 69.7   & 28.0& 24.0 & 33.5 & 29.2 \\
        FedBN\cite{li2021fedbn}+D        & ICLR'21     & 47.9 & 43.5& 72.3 & 68.2  & 25.2 & 21.2   & 31.1 & 26.5 \\  \hdashline
        SKA    & Ours    & \textbf{53.9} & \textbf{49.8}  & \textbf{76.0}  & \textbf{72.7} & \textbf{36.7}& \textbf{32.2} & \textbf{49.7} & \textbf{45.0}   \\ \hline
        \end{tabular}%
        }
        \caption{Comparisons of decentralised learning Re-ID on four unseen novel domains.
        }
        \label{tab:sota_small}
    
    \end{center}
    \end{table}

\subsection{Comparisons with the State-of-the-Art Decentralised
  Methods}
\label{section_SOTA}
We compared the proposed SKA with two state-of-the-art decentralised person Re-ID methods, namely
FedReID~\cite{wu2021decentralised} and FedPav~\cite{zhuang2020performance},
and four state-of-the-art federated learning methods,
namely FedAvg~\cite{mcmahan2017communication}, FedProx\cite{li2018federated}, FedBN~\cite{li2021fedbn} and MOON~\cite{li2021model}.
Note that the conventional federated learning methods cannot be directly used for decentralised person Re-ID,
so we implement them based on the federated Re-ID backbone model~\cite{zhuang2020performance},
where domain-specific classifiers are not used for central aggregation.
Table~\ref{tab:sota_large} gives results on source client domains
and Table~\ref{tab:sota_small} for results on unseen novel domains.

Table~\ref{tab:sota_large} shows that the proposed SKA has
significantly better Re-ID performance than all other models on individual client domains.
Specifically, state-of-the-art decentralised Re-ID methods
(FedReID~\cite{wu2021decentralised} and FedPav~\cite{zhuang2020performance})
perform slightly worse on average than the baseline local supervised model
which trains the backbone model on each local domain in a conventional centralised learning way. 
In contrast, SKA significantly improves over the baseline local supervised
model by 5\%$\sim$10\% in R1 and 10+\% in mAP on all test datasets.
Compared with the state-of-the-art federated learning methods
(FedAvg~\cite{mcmahan2017communication}, FedProx~\cite{li2018federated}, FedBN~\cite{li2021fedbn} and MOON~\cite{li2021model}),
SKA performs better than FedBN in R1 by 4.98\% and mAP by 5.9\% on average,
and significantly outperforms FedAvg in R1 by 17.05\% and mAP by 19.3\% on average.
Table~\ref{tab:sota_small} also shows that the proposed SKA achieves
notable generalisation performance improvement
on unseen novel domains by outperforming the second best model in R1 by 3.6\%
(VIPeR), 1.6\%(iLIDS), 7.4\%(GRID), 13.1\% (PRID) respectively, and in
mAP by 4.3\% (VIPeR), 1.1\%(iLIDS), 7.6\%(GRID), 12.7\% (PRID)
respectively. 
These results show a very clear superiority of the proposed SKA model over the state-of-the-art methods.
Besides, although centralised and decentralised methods are not directly comparable,
the proposed SKA method performs closely to some centralised generalisable Re-ID methods.
For example, compared with the state-of-the-art centralised SNR~\cite{jin2020style} method,
on Duke, SKA yields 66.6\% vs. SNR yields 73.2\% in terms of mAP;
on VIPeR, SKA yields 53.9\% vs. SNR yields 55.1\% in terms of mAP.

\begin{table}[t]
  \small    
\begin{center}
    \resizebox{0.99\textwidth}{!}{%
    \begin{tabular}{l|cc|cc|cc|cc}
    \hline
    \multirow{3}{*}{Components}
     &\multicolumn{4}{c|}{\textbf{Seen domains}} &
      \multicolumn{4}{c}{\textbf{Unseen domains}}
      \\\cline{2-9}
      &\multicolumn{2}{c|}{\textbf{Duke}} &
      \multicolumn{2}{c|}{\textbf{ Market}} &
      \multicolumn{2}{c|}{\textbf{ VIPeR}} &
      \multicolumn{2}{c}{\textbf{ iLIDS}}\\ 
    &mAP  &R1 &mAP &R1 &mAP &R1 &mAP &R1  \\\hline
    Baseline  &50.9 & 70.8        & 56.6    & 81.3      &48.2    & 44.3      & 73.3     & 69.3 \\
    Baseline+AN      &58.7   & 78.1       &64.2    & 87.9      &54.2     & 50.2     & 77.0     & 74.0 \\
    Baseline+LSN    &62.8${}^\dagger$  & 80.0${}^\dagger$       &73.1${}^\dagger$   & 90.4${}^\dagger$          &47.9    & 43.5     &72.3   & 68.2 \\
    Baseline+LSN+Dual  & 62.8${}^\dagger$ & 80.0${}^\dagger$ & 
    73.1${}^\dagger$ & 90.4${}^\dagger$  & 48.8 & 44.6  & 75.2 & 71.3\\
    Baseline+LSN+AN & 66.6* & 83.7* & 78.2* & 92.7*  &51.8  & 47.8 & 73.1 & 69.3 \\
    Baseline+LSN+AN+Dual & 66.6* & 83.7* &78.2* & 92.7* &53.9 & 49.8 & 76.0 & 72.7 \\
    \hline
    \end{tabular}%
    }
    \caption{
    Evaluating component effectiveness on seen source domains and unseen novel domains.
    *,${}^\dagger$ `Dual' does not affect model personalisation on seen source domains
    as local generalised normalisation layers are only learnt to construct the global generalised model.
    }
    \label{tab:ablation_component}
\end{center}
\end{table}

\subsection{Ablation Studies}
\label{section_ablation}
\noindent\textbf{Component Effectiveness Evaluation}.
In Table~\ref{tab:ablation_component},
`Baseline' means the federated Re-ID learning baseline model~\cite{wu2021decentralised,zhuang2020performance},
`AN' means incorporating Attentive Normalisation into the vanilla BN layers in the feature extraction model,
`LSN' means local specific vanilla BN normalisation,
`Dual' means the dual local normalisation mechanism.
On unseen novel domains, `Baseline+LSN' and `Baseline+LSN+AN' are by averaging the normalisation layers at the
end of model training to construct the global generalised model.
Table~\ref{tab:ablation_component} shows:
(1) LSN can significantly improve model personalisation performance on seen source domains.
However, it degrades clearly model generalisation performance on unseen novel domains;
(2) AN can improve model personalisation and generalisation performance on both seen and unseen domains;
(3) Learning local personalisation knowledge with LSN+AN in SKA brings the best model
personalisation performance on seen domains;
(4) The dual local normalisation mechanism can compensate the degradation brought by
learning local personalisation knowledge, resulting in better global model generalisation.
It improves `Baseline+LSN' by approximately 1\% and 2\% on VIPeR and iLIDS respectively,
while improves `Baseline+LSN+AN' by approximately 2\% on both VIPeR and iLIDS.
And we can see that, on unseen domains, `Baseline+LSN+AN+Dual' performs closely to
`Baseline+AN' which does not consider learning local personalisation knowledge,
while on seen source domains, `Baseline+LSN+AN+Dual' performs significantly better than `Baseline+AN'.
This further verifies that `Dual' can help SKA to optimise the trade-off between model personalisation and generalisation.

\begin{table}[t]
  \small	
\begin{minipage}{0.48\textwidth}
  \begin{center}
    \resizebox{0.88\textwidth}{!}{%
  \begin{tabular}{l|cc}
    \hline
    Variants & \textbf{Duke} & \textbf{Market} \\
    \hline
    SKA & 83.7 & 92.7\\
    SKA w/o LSN &  78.1 & 87.9\\
    SKA w/o AN & 80.0 & 90.4\\
    SKA w/o AN + SE & 82.0 & 91.2\\
    SKA w/o AN + CBAM & 80.5 & 91.1\\
    \hline
    \end{tabular}
    }
\caption{Evaluating variants of learning local personalisation knowledge (R1).}
  \label{tab:ablation_normalisation}
\end{center}
\end{minipage}
\hspace{0.02\textwidth}
\begin{minipage}{0.48\textwidth}
  \begin{center}
      \resizebox{\textwidth}{!}{%
  \begin{tabular}{l|cc}
    \hline
    Variants & \textbf{VIPeR} & \textbf{iLIDS} \\
    \hline
    SKA w/ Dual& 49.8 & 72.7\\
    SKA w/o Dual + Avg & 47.8 & 69.3 \\
    SKA w/o Dual + FeatConcat & 47.2 & 71.8\\
    SKA w/o Dual + RandImg & 6.6 & 21.8 \\
    \hline
    \end{tabular}
    }
  \caption{Evaluating variants of dual local normalisation (R1).}
  \label{tab:ablation_dual}
\end{center}
\end{minipage}
\end{table}

\vspace{0.1cm}    
\noindent\textbf{Evaluating Variants of Learning Local Personalisation Knowledge}.
In Table~\ref{tab:ablation_normalisation}, we tested some variants of learning local personalisation knowledge.
`SE' means using the SE~\cite{hu2018squeeze} unit in the backbone model
and `CBAM' means using the CBAM~\cite{Woo_2018_ECCV} unit in the backbone model.
Table~\ref{tab:ablation_normalisation} shows that learning local specific normalisation
and attentive knowledge facilitates better model personalisation.
Overall, SKA performs better than other variants.

\vspace{0.1cm}    
\noindent\textbf{Evaluating Variants of Dual Local Normalisation}.
In Table~\ref{tab:ablation_dual},we tested different strategies to construct normalisation layers in the global generalised model.
`Avg' means averaging local specific normalisation layers at the end of model training.
`FeatConcat' means using each local specific normalisation layers to extract features and concatenating these features
for person matching.
`RandImg' means using ImageNet pretrained weights to re-initialise normalisation layers.
Table~\ref{tab:ablation_dual} shows that the proposed dual local normalisation mechanism
performs better than other variants for constructing a global generalised model.

\section{Conclusions}
In this work, we proposed a new Selective Knowledge Aggregation (SKA) approach to decentralised person Re-ID.
The key idea is to learn local batch and attentive normalisation layers specific to each domain
for improving model personalisation in Re-ID on seen client domains,
and to use a dual local normalisation mechanism for improving model
generalisation in Re-ID on unseen novel domains.
Extensive experiments on eight person Re-ID datasets show the superiority of the proposed SKA approach
over the state-of-the-art decentralised methods on both seen source client domains and unseen novel domains.

\bibliography{ska_bmvc}
\end{document}